\documentclass[11pt,a4paper]{article}
\usepackage[hyperref]{acl2021}
\usepackage{times}
\usepackage{latexsym}

\newcommand{\remove}[1]{}
\newcommand{\xhdr}[1]{\vspace{1mm}\noindent{{\bf #1.}}} 

\usepackage{xcolor}
\usepackage{graphicx}

\usepackage{microtype}
\graphicspath{ {./images/} }

\usepackage{paralist}
\aclfinalcopy 

\usepackage{epstopdf}
\usepackage{hyperref}

\title{{Catchphrase}: Automatic Detection of Cultural References}

\author{Nir Sweed \\
  The Hebrew University of Jerusalem \\
  \texttt{nir.sweed@mail.huji.ac.il} \\\And
  Dafna Shahaf \\
The Hebrew University of Jerusalem \\
  \texttt{dshahaf@cs.huji.ac.il} \\}
\date{}

\usepackage{footmisc}

\begin{document}
\maketitle
\begin{abstract}
A \emph{snowclone} is a customizable phrasal template that can be realized in multiple, instantly recognized variants. For example, ``* is the new *" (Orange is the new black, 40 is the new 30). Snowclones are extensively used in social media. In this paper, we study snowclones originating from pop-culture quotes; our goal is to automatically detect cultural references in text.
We introduce a new, publicly available data set of pop-culture quotes and their corresponding snowclone usages and train models on them. We publish code for \textsc{Catchphrase}, an internet browser plugin to automatically detect and mark references in real-time, and examine its performance via a user study. Aside from assisting people to better comprehend cultural references, we hope that detecting snowclones can complement work on paraphrasing and help to tackle long-standing questions in social science about the dynamics of information propagation.
\end{abstract}

\section{Introduction}
\label{sec:intro}

\begin{figure}[b!]
    \centering
    \includegraphics[width=0.75\linewidth]{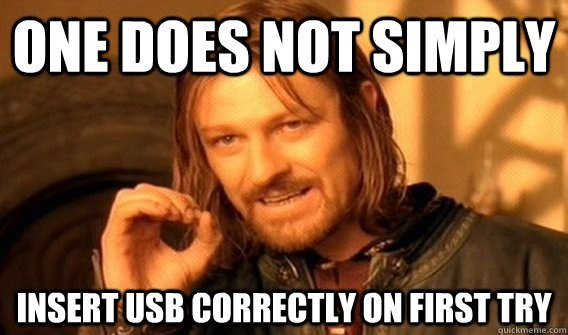}
    \caption{Snowclone example, based on ``One does not simply walk into Mordor'' from ``Lord of the Rings''.}
    \label{fig:onedoesnot}
\end{figure}

First coined by Richard Dawkins \cite{dawkins2016selfish:1976}, a meme is a unit of cultural transmission: any idea
or behavior that can be transferred by imitation. Internet memes have become an integral part of modern digital culture \citep{shifman2014memes}. Pullum \citep{pullum2004} coined the term \emph{snowclones} to describe a specific type of meme -- phrasal templates that are easily reusable in many different contexts.  Pullum described a snowclone as {``a multi-use, customizable, instantly recognizable, time-worn, quoted or misquoted phrase or sentence that can be used in an entirely open array of different jokey variants''.}  For example, the quote ``One does not simply walk into Mordor'' from the ``Lord of the Rings'' films became a well-known pattern -- \emph{``One does not simply *"} --  used extensively online (see Figure \ref{fig:onedoesnot}).

In this paper, our goal is to develop algorithms to {\bf detect snowclones} in text. We envision an ``Englishman in New York'' -- a foreigner, perhaps, or someone who does not easily understand contemporary cultural references and could use the help of an automated system to communicate better. In particular, we focus on pop-culture references over the internet.

\remove{
Perhaps the closest previous work is the NIFTY and MEMETRACKER systems \citep{suen2013nifty,leskovec2009meme}, that tracked quotations attributed to individuals, with the goal of analyzing the online news cycle. 
These works focused on short, distinctive phrases that travel \emph{relatively intact} through on-line text (and indeed, their distance measure is based solely on the substring edit-distance. Therefore, this setting is much more restrictive than ours.

    Two additional related tasks are multi-word expression/idiom identification \citep{haagsma-etal-2020-magpie, zarriess2009exploiting, muzny2013automatic} and clich\'e detection \citep{cook2013automatically,van-cranenburgh-2018-cliche}. Idioms and multi-word expressions are chiefly fixed expressions, thus recognition systems usually consider only minor variations. For instance, we expect ``Jumped the shark'' to be recognized as a minor variation of the idiom ``Jump the shark''. However, replacing even a single word (``Jump the bed'', ``Feed the shark'') makes for a non-idiomatic phrase. Similarly, while some clich\'es do follow a snowclone pattern, most clich\'es do not allow many variations, and their meaning is kept across usages. For example, switching even a single word of the clich\'e ``cat got your tongue?'' results in a non-clich\'e expression. In contrast, snowclones maintain some structure of their source but completely change its meaning.
    }

From a linguistic point of view, snowclones complement the paraphrasing task \citep{barzilay2001extracting,fernando2008semantic,dolan2004unsupervised}.
Paraphrase detection identifies alternative ways to convey the
same meaning, while snowclones keep (some of) the original sentence structure but completely change the meaning. 

Detection and tracking of digital memes have been the focus of multiple computational studies. The closest to our work are 
MEMETRACKER and NIFTY \citep{leskovec2009meme,suen2013nifty}, that tracked quotations attributed to individuals. 
These works focused on short, distinctive phrases that travel \emph{relatively intact} through on-line text.
Other related tasks are multi-word expression/idiom identification \citep{haagsma-etal-2020-magpie, zarriess2009exploiting, muzny2013automatic} and clich\'e detection \citep{cook2013automatically,van-cranenburgh-2018-cliche}. Again, idioms and multi-word expressions are chiefly fixed expressions (``cat got your tongue?'', ``jumped the shark'') that rarely change their meaning across mutations.  Therefore, these settings are much more restrictive than ours.

Our contributions are the following: we propose a novel task of snowclone detection, identifying cultural references. We first formulate it as a tagging task, treating snowclones as regular expressions; we conduct a user study to show humans have an intuitive notion of the ``correct'' pattern(s), and develop a sequence-to-sequence tagger to reveal such patterns. We then extend the formulation to softer notions of similarity. We experiment with feature-based and neural approaches, achieving high accuracies. To further show the utility of our methods, we develop \textsc{Catchphrase}, a browser extension to detect pop-culture references, conduct a user study and show it indeed helps users identify cultural references. We publish data and code \footnotemark[1]. We believe tracking snowclones will find interesting applications in social science, exploring the diffusion and evolution of highly dynamic content online.
\footnotetext[1]{\url{https://github.com/sweedy12/CATCHPHRASE}}

\remove{
\begin{compactitem}
\item  
    \item We develop a sequence-to-sequence tagger, aimed at revealing potential underlying snowclone patterns. 
    \item 
    \item 
\end{compactitem}
}

\section{Snowclones as Regular Expressions}
\label{sec:snowclone form tagger}

The common view of snowclones treats them as regular expressions \citep*{snowclonesdb}. Thus, in this section, we formulate the snowclone detection problem as a tagging task. Intuitively, we want to predict for each word in the original sentence whether it is replaced by a wildcard. We use the resulting pattern to match new sentences to the original sentence. 
For example, given a sentence $s= \langle One,does,not,simply,walk,into,Mordor\rangle$ 
we would like to find a mapping: \\ $T(s) = \langle One, does,not,simply,*,*,* \rangle .$ (Adjacent wildcards can be merged) 



\subsection{Can People do This?}
\label{snowclone tagging task}

Before we set out to find an algorithm to uncover the underlying snowclone form of an input sentence, we try to evaluate the feasibility of this task. It is not clear that such patterns exist, or are agreed upon by human annotators. 
To that end, we conduct a user study to test if people have an intuitive notion of snowclone patterns. 

\begin{figure}
\centering
    \includegraphics[width=0.85\linewidth]{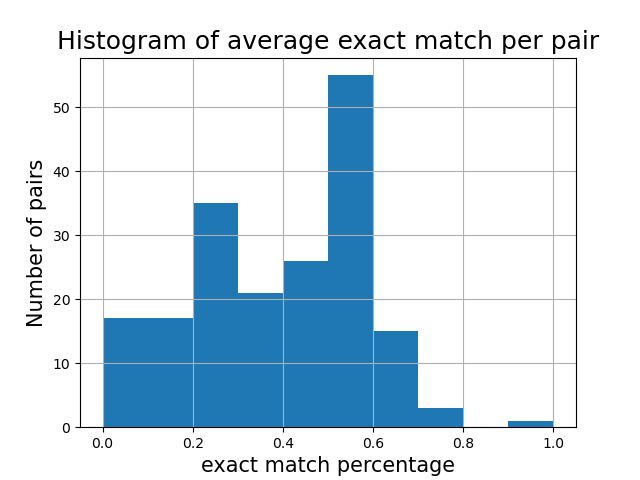}
    \includegraphics[width=0.85\linewidth]{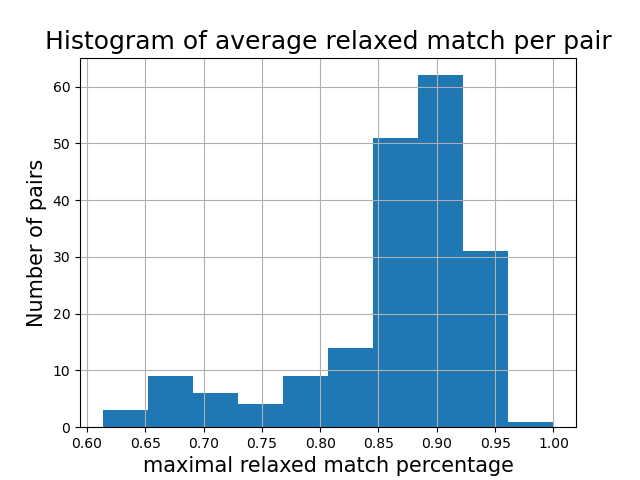}
    \caption{Histogram of the exact-match similarity measure (top) and relaxed-match measure (bottom), averaged over all sentences, for all pairs of participants. 
    }
    \label{fig:phrasal-template-figs}
\end{figure}

We recruited 22 volunteers through social media. The participants were 80\% males. 85\% of them were 25-35 years old, the rest being 40-55. All participants were Israeli and identified as non-native English speakers. Participants were given a short explanation of snowclones and instructed to find the snowclone form of a set of (the same) 20 sentences, chosen from the memorable movie quote database  \citep{Danescu-Niculescu-Mizil+al:12b}. We chose sentences at random, filtering out quotes that became known internet memes.  
Participants marked words that should become wildcards, generating up to 3 patterns per sentence, as they saw fit.  We asked participants to report whether they were familiar with any sentence, and discarded the entire questionnaire of two who did.

\xhdr{Evaluation and Results}
To compute similarity between pairs of people, we propose two measures. 
In \textbf{exact-match}, the score is the percentage of sentences (out of 20) on which the two people had at least one exact-match pattern.
In \textbf{relaxed-match}, we compute for each sentence the closest match between the patterns of both people (in terms of simple \% agreement). The score is the percentage of agreement over all 20 closest matches.

 Figure \ref{fig:phrasal-template-figs} shows histograms over all pairs of participants. For exact-match, most pairs of participants agree on roughly half the patterns.  A careful examination of the results indicates that participants are divided into those that prefer a single, general pattern (annotating ``The pavement was his enemy" as ``The * was his *"), and those preferring several narrower patterns (marking both ``The * was his enemy" and ``The pavement was his *"). Another contributing factor is that many mismatched pairs of patterns differ only in stopwords. The relaxed-match measure is less sensitive to this issue and indeed demonstrates high agreement. We hypothesize that this indicates the feasibility of training machine learning models for this task.

\section{Snowclone Pattern Tagger}
\label{sec:Snowclone tagger}
We create and publish our own data set for this task, and use it to train two different ML models for it.

 \xhdr{Data}
To train ML models to solve the task of snowclone tagging, we needed examples for sentences and their underlying snowclone form. To this end, we use the snowclone patterns along with the original quotes from 
The Snowclone Database \citep*{snowclonesdb}. As this is not enough data to train on, we use the patterns to lookup more instances online, collecting 7700 $\langle\textit{snowclone pattern, instance}\rangle$ pairs. 
When splitting to train-dev-test sets (60\%/20\%/20\%), we make sure all variants of the same pattern are put in the same set. We release our dataset\footnotemark[1]

\xhdr{Bi-LSTM-CRF} We adapt the model of \citep{huang2015bidirectional:2015}, tested on part of speech tagging, chunking and named entity recognition tasks. 
Its CRF layer performs a structured prediction over the sentence tags, using sentence-level information rather than predicting a label for each word separately, rendering it useful for our task. %
For optimization, we use negative log-likelihood. 

\xhdr{BERT S2S} We use BERT \cite{devlin-etal-2019-bert}, 
 as it has shown to produce good results when fine-tuned to specific sequence-to-sequence tasks. We fine-tune BERT for a token classification task using the snowclone form dataset.  Since this model outputs a probability measure for each token, we use binary cross-entropy as the objective function.

\begin{table}
\begin{tabular}{ |p{3cm}||p{1.5cm}|p{1.5cm}|  }
 \hline
 \multicolumn{3}{|c|}{Snowclone Form Tagging - Results} \\
 \hline
  Model&Accuracy&Recall\\
 \hline
 Naive   & 0.74    &0\\
 Bi-LSTM-CRF&   {\bf 0.92}  &  0.82\\
 BERT &0.9 & {\bf0.88}\\
 \hline
\end{tabular}
\caption{Accuracy and recall for each of the proposed models for the snowclone form tagging task.}
\label{tables: s2s results}
\end{table}

See Appendix \ref{sec:taggerhyper} for implementation details and hyper-parameter tuning.

\remove{
\xhdr{Hyper-parameter tuning}
For the BI-LSTM-CRF model, we perform a small grid search to determine the values for the learning rate, weight decay, and the number of layers and hidden dimension of the the BI-LSTM. As the search space, we used $learning-rate \in \{0.01, 0.001, 0.0001\}$, $\textit{weight-decay}\in\{0,0.01,0.001 \}$, $\textit{num-layers}\in\{1,2,3\}$ and $\textit{hidden-dim}\in\{32,64,128\}$. Finally, we choose $\textit{learning-rate} = 0.01$,$\textit{weight-decay} = 0$, $\textit{num-layers} = 2$,$\textit{hidden-dim} = 32$.  For the BERT model, we use a smaller grid search over the learning rate ($\in\{0.001,0.0001\}$) and the weight decay ($\{0,0.01,0.001\}$) hyper-parameters, and train it for a single epoch using $\textit{learning-rate} = 0.0001$,$\textit{weight-decay} = 0.01$.
}

\xhdr{Evaluation and results}
 Since most words in an input sentence are not replaceable, wildcards are infrequent. Thus, we prioritize models with higher recall than precision. Table \ref{tables: s2s results} shows recall and accuracy of the models. The naive majority baseline (no words are wildcards) yields 74\% accuracy (and, naturally, 0\% recall).  The Bi-LSTM-CRF model reaches an accuracy of 92\%, and 82\% recall. BERT achieves an accuracy of 90\%, and recall 88\%. 
\section{Going Beyond Regular Expressions}
\label{sec:snowclone usage detection}

When we tried to apply our models to find snowclones in online community text (looking for regex matches), we realized that the regex formulation might be too simplistic, as some cultural references do not follow the snowclone pattern exactly, and some sentences that do follow it are not really references. Take Apocalypse Now's famous ``I love the smell of napalm in the morning''. A natural corresponding pattern is ``I love the smell of * in the morning'', and indeed, ``I love the smell of bureaucracy in the morning'' is most likely a reference to the movie. However, the case of ``I love the smell of \emph{pancakes} in the morning'' is a lot less clear. On the other hand, ``30 is the old 40'' does not perfectly match the ``* is the new *'' pattern, but still might be considered a reference.
%
In this section we reformulate the problem, using the output of the sequence-to-sequence tagger as one input to a machine learning model. 

We reformulate the problem as a binary classification task over pairs of sentences. Given a seed sentence  $s$ representing an original pop-culture quote, and a candidate sentence $c$, decide whether $c$ is a reference to $s$.
%
We note this is not an easy task, as it is hard to put our finger on why  ``One does not simply forget to social distance" is likely a reference to ``One does not simply walk into Mordor", but  ``One cannot just walk right into jail" is not. 

\section{Snowclone Reference Detector}
\label{sec:Snowclone reference detector}

 \xhdr{Data} We searched the web and found 20 famous movie quotes that turned into snowclone internet memes. We removed three quotes appearing in the data of Section \ref{sec:Snowclone tagger}, not to contaminate our evaluation. Next, we defined overly general regular expressions for each seed (attempting to catch both snowclones and not) and crawled Reddit conversations to find matches.  
We choose Reddit due to its  popularity and  comprehensive use of  memes. 
We collected 3850 pairs of seed and sentence and had an expert manually annotate them (after calibration). The dataset is imbalanced, with 64\% of pairs tagged as non-reference.
When splitting to train-dev-test (60\%/20\%/20\%), we ensure all examples from the same seed are put in the same set.
 We take a supervised approach and train two models. 
 
\xhdr{Feature-based SVM model}
    We calculate three sets of features, focusing on sentence \emph{structure}. (1) Similarity between $s$ and $c$: edit distance, longest common sequence, and longest substring between $s,c$. (2)  We use the snowclone tagger of Section \ref{sec:Snowclone tagger} to predict $\hat{s}$, the snowclone form of $s$ and use the same features of group (1) between $\hat{s},c$. (3) To characterize the shared and replaced words we calculate the idf statistic for words shared between $s,c$ and words in $s$ but not in $c$ (idf over  movie quotes \citep{Danescu-Niculescu-Mizil+Lee:11a}).  We tried  decision trees, random forests, and SVM, and chose SVM due to its performance. 

\xhdr{RoBERTa-based model} 
We chose RoBERTa as our second model, as it showed impressive results on a related 2-sentence classification task. We use a model pre-trained on SNLI \citep{nie-etal-2020-adversarial}, which achieved state-of-the-art result on a natural language inference task. We replace its classification head with a binary classification head, and fine-tune the model on the dataset of Section \ref{sec:snowclone usage detection}. Unlike SVM, we expect this model to capture semantic similarity (e.g., between ``old'' and ``new'').

 \begin{table}
\tabcolsep = 0.01cm
\begin{tabular}{ |p{2cm}||p{1.8cm}|p{1.9cm}|p{1.8cm}|  }
 \hline
 \multicolumn{4}{|c|}{Snowclone Detection - Results} \\
 \hline
  Model&Accuracy&Precision&Recall\\
 \hline
 Naive   & 0.64    &1&0\\
 SVM&   \textbf{0.85$\pm$0.08} \  &\textbf{ 0.84$\pm$0.13}&\textbf{0.78$\pm$0.12}\\

 RoBERTa    &0.81$\pm$0.94 & 0.7$\pm$ 0.15&0.74$\pm$0.18\\
 \hline
\end{tabular}
\caption{Snowclone detection task. We performed 20 splits for the SVM model and 5 for RoBERTa, and report standard deviation.}
\label{tables: snowclone detection results}
\end{table}

\begin{figure*}[t!]
    \centering
    \includegraphics[width=0.7\linewidth]{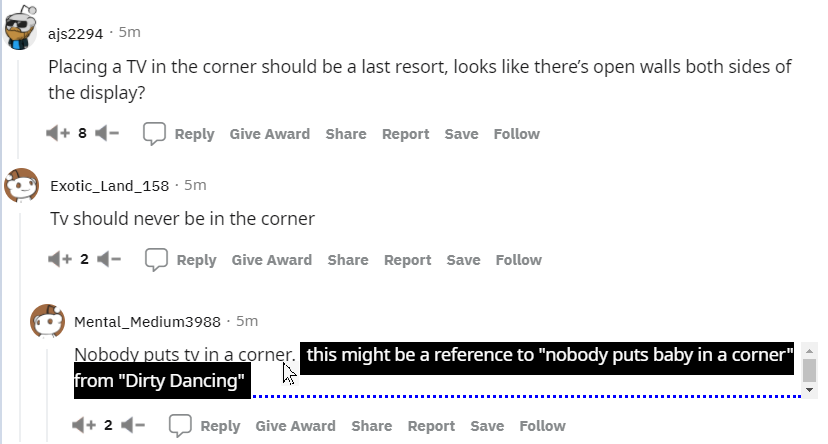}
    \caption{Screenshot of our web extension, suggesting ``Nobody puts TV in a corner" is a reference to Dirty Dancing's ``Nobody puts baby in a corner". The suggested reference is underlined in blue. Hovering over the underlined sentence prompts a message containing original quote information.}
    \label{fig:chrome_extension_figs}
\end{figure*}

See Appendix \ref{sec:refhyper} for implementation details and hyper-parameter tuning.

\xhdr{Evaluation and results}
The accuracy, precision and recall measures for all models are presented in Table \ref{tables: snowclone detection results}. The naive majority baseline achieves 64\% accuracy on the full data set (as the data is not balanced). For our feature-based SVM model, we randomly select 20 different splits, reaching an average of 85\% accuracy, 84\% precision and 78\% recall, with a corresponding std of 8.7\%, 13.8\% and 12\%. The RoBERTa-based model achieved average results of 81\% accuracy, 70\% precision and 74\% recall, with std 9.4\%, 15.7\% and 18.7\%. Thus, we chose the SVM model. This perhaps indicates the importance of structure in the snowclone problem; alternatively, perhaps the amount of data was not sufficient to fine-tune RoBERTa.

\xhdr{Observations} As a (qualitative) reality check, we choose 10 seeds unseen during training. We crawl all Reddit posts from March 2016 (month and year chosen at random). We choose Reddit as a diverse and popular online community, where internet memes are used regularly. We use the SVM model to collect new candidate references for the seeds.
We analyze the candidate references and observe that (not surprisingly) their quality is heavily influenced by the snowclone tagger feature. When the regex is too general (e.g., ``I am your *" for ``I am your father"), the number of false positives is high. 
Importantly, over all seeds our method is capable of detecting true references that do not exactly match the predicted snowclone-form.


\section{Evaluation: Web Extension}
\label{web extension experiment}

Our main motivation in this study was to help the proverbial ``Englishman in New York'' identify cultural references. 
In this section, we ask whether  \emph{our algorithms can help users detect pop-culture references online}. We create \textsc{Catchphrase}, a web browser extension able to detect and mark pop-culture references in web pages (see Figure \ref{fig:chrome_extension_figs}). The extension inspects the web page source and identifies candidate sentences. We use locality-sensitive hashing \citep{gionis1999similarity} with similarity threshold = 0.2 for filtering, allowing us to reduce computation time and maintain a small number of false negatives. Next, the extension runs the reference-detector on each (seed, candidate sentence) pair and highlights the predicted references.

\xhdr{Experimental design}
 We choose a set of 20 pop-culture quotes (seeds) unseen by our reference-detector during training time, and whose snowclone form is the basis to many variations. All sentences chosen are ones that became popular internet memes. For each seed quote, we manually crawled Reddit and found threads containing references to it. After filtering out threads that were over 10 messages long, we were left with 106 threads. 
 
 We recruited 10 volunteers  through social media, all Israeli, non-native English speakers, who self-identified as having low familiarity with pop-culture. 80\% of the volunteers were 20-35 years old, and the remaining 20\% were 40-60 years old. 70\% of the participants were females. We randomly selected 16 seeds for each participant and randomly split them into two groups, one per condition (with and without our extension).
 The threads were shown in random order.
 The participants were asked to go over each thread and point out any pop-culture references they detect, specifying their origin if they knew it. 
  

\xhdr{Evaluation and results}
    Under the no-extension condition, participants correctly identified a pop-culture reference 38.7\% of the time. The reference origin was correctly identified in 61.2\% of these. This is interesting, as it shows people can identify that a sentence \emph{looks} like a cultural reference, even when they do not recognize the source. 
    
    When using the extension, participants correctly identified a reference 68.7\% of the times, recognizing the origin in 98.1\% of these. In 26.3\% of the threads, the algorithm did not recognize the reference. 5\% of the times, we believe the algorithm was right but people thought it was not (e.g., ``I solemnly swear I'm up for good tea'' as a reference to ``I solemnly swear I'm up to no good''). The reason source recognition is not perfect is one user finding a sentence the algorithm missed (but not attributing it). To check our hypothesis that web-extension users recognize more pop-culture references, we run t-test with $\alpha=0.95$ and reject the null hypothesis with $pval = 0.00005$.

\section{Conclusions and Future Work}
In this work we proposed the novel task of detecting snowclones in text. Motivated by the high agreement achieved by humans on a snowclone annotation task, we first developed algorithms for finding snowclones which are regular expressions, 
then extended the formulation to a softer notion of similarity.
We introduce a new data set of pop-culture quotes and their corresponding snowclone variants and train models on them. We publish code for \textsc{Catchphrase}, an internet browser plugin to automatically detect and mark references in real-time. Our results demonstrate our algorithms can indeed help users detect pop-culture references.

In the future, our work might be used in conversational AI context, supporting agents' ability to understand and even generate pop-culture references. Another direction worth pursuing is applying our methods to domains outside pop-culture (or at the very least, to pop-culture of different cultures). 

We believe snowclones, complementing the notion of paraphrases, are worth exploring  and can give us new insights into how ideas spread and evolve. Our approach opens an opportunity to better answer long-standing questions in social science about the dynamics of information. 





\section*{Acknowledgements}
We thank the anonymous reviewers for their insightful and highly constructive comments and
Roy Schwartz for stimulating discussions.
This work was supported by the European Research Council (ERC) under the European Union's Horizon 2020 research and innovation programme (grant no. 852686, SIAM), US National
Science Foundation, US-Israel Binational Science Foundation (NSF-BSF) grant no. 2017741, and Amazon Research Awards.

\bibliography{anthology,acl2021}
\bibliographystyle{acl_natbib}

\newpage 

\appendix

Below we provide implementation details for the sake of reproducibility. 
\section{Snowclone Pattern Tagger: Hyper-parameter tuning}
\label{sec:taggerhyper}

For the BI-LSTM-CRF model, we perform a small grid search to determine the values for the learning rate, weight decay, and the number of layers and hidden dimension of the the BI-LSTM. As the search space, we used $learning-rate \in \{0.01, 0.001, 0.0001\}$, $\textit{weight-decay}\in\{0,0.01,0.001 \}$, $\textit{num-layers}\in\{1,2,3\}$ and $\textit{hidden-dim}\in\{32,64,128\}$. Finally, we choose $\textit{learning-rate} = 0.01$,$\textit{weight-decay} = 0$, $\textit{num-layers} = 2$,$\textit{hidden-dim} = 32$.  For the BERT model, we use a smaller grid search over the learning rate ($\in\{0.001,0.0001\}$) and the weight decay ($\{0,0.01,0.001\}$) hyper-parameters, and train it for a single epoch using $\textit{learning-rate} = 0.0001$,$\textit{weight-decay} = 0.01$.

\section{Snowclone Reference Detector: Hyper-parameter tuning}
\label{sec:refhyper}
For the RoBERTa model, we perform the same hyper-parameter search as described in Section \ref{sec:taggerhyper}, and use the same values. For the SVM model, we search over kernels (RBF, linear and polynomial), degree (when applicable, over $[2,3,4]$) and C-values ($\{0.1\cdot i\}_{i=1}^{10}$). Our search dictates using a polynomial kernel of degree 3, with $C = 0.5$.

\end{document}